\documentclass[10pt,twocolumn,letterpaper]{article}

\usepackage{wacv}
\usepackage{times}
\usepackage{epsfig}
\usepackage{graphicx}
\usepackage{amsmath}
\usepackage{amssymb,bbm,diagbox}

\def\wacvPaperID{583} 

\wacvfinalcopy 
\ifwacvfinal
\def\assignedStartPage{1} 
\fi

\ifwacvfinal
\usepackage[breaklinks=true,bookmarks=false]{hyperref}
\else
\usepackage[pagebackref=true,breaklinks=true,colorlinks,bookmarks=false]{hyperref}
\fi

\ifwacvfinal
\else
\fi

\begin{document}

\title{End-to-end Lane Shape Prediction with Transformers}

\author {
    Ruijin Liu\textsuperscript{\rm 1}\quad Zejian Yuan\textsuperscript{\rm 1}\quad Tie Liu\textsuperscript{\rm 2}\quad Zhiliang Xiong\textsuperscript{\rm 3} \\
    \\
    \textsuperscript{\rm 1}{Institute of Artificial Intelligence and Robotics, Xi'an Jiaotong University, China}\\
    \textsuperscript{\rm 2}{College of Information Engineering, Capital Normal University, China}\\
    \textsuperscript{\rm 3}{Shenzhen Forward Innovation Digital Technology Co. Ltd, China}\\
    \textsuperscript{\rm 1}{\tt\small lrj466097290@stu.xjtu.edu.cn} \quad \textsuperscript{\rm 1}{\tt\small yuan.ze.jian@xjtu.edu.cn} \\
    \textsuperscript{\rm 2}{\tt\small liutiel@163.com}\quad \textsuperscript{\rm 3}{\tt\small leslie.xiong@forward-innovation.com}\\
}

\maketitle
\thispagestyle{empty}

\begin{abstract}
Lane detection, the process of identifying lane markings as approximated curves, is widely used for lane departure warning and adaptive cruise control in autonomous vehicles.
The popular pipeline that solves it in two steps---feature extraction plus post-processing, while useful, is too inefficient and flawed in learning the global context and lanes' long and thin structures.
To tackle these issues, we propose an end-to-end method that directly outputs parameters of a lane shape model, using a network built with a transformer to learn richer structures and context. The lane shape model is formulated based on road structures and camera pose, providing physical interpretation for parameters of network output. 
The transformer models non-local interactions with a self-attention mechanism to capture slender structures and global context.
The proposed method is validated on the TuSimple benchmark and shows state-of-the-art accuracy with the most lightweight model size and fastest speed. Additionally, our method shows excellent adaptability to a challenging self-collected lane detection dataset, showing its powerful deployment potential in real applications. Codes are available at \url{https://github.com/liuruijin17/LSTR}.
\end{abstract}

\section{Introduction}

Vision-based lane marking detection is a fundamental module in autonomous driving, which has achieved remarkable performance on applications such as lane departure warning, adaptive cruise control, and traffic understanding~\cite{LDWReview,LaneFeatureMethod,LaneModelMethod,SCNN}. 
In real applications, detecting lanes could be very challenging. The lane marking is a long and thin structure with strong shape prior but few appearance clues~\cite{SCNN}. Besides, lane markings vary in different types, light changes, and occlusions of vehicles and pedestrians, which requires the global context information to infer the vacancy or occluded part. Moreover, the high running efficiency and transferring adaptability of algorithms are indispensable for deployment on mobile devices~\cite{FastDraw}.

Existing methods~\cite{LaneNet,SCNN,ENet-SAD,FastDraw,PINet} take advantage of the powerful representation capabilities of convolution neural networks (CNNs) to improve the performance of lane detection task by a large margin over traditional methods~\cite{LDWReview,LaneFeatureMethod,LaneModelMethod} which are based on hand-crafted features and Hough Transform.
However, current CNNs-based methods still are flawed in addressing the aforementioned challenges. The earlier methods~\cite{LaneNet,FastDraw} typically first generate segmentation results and then employ post-processing, such as segment clustering and curve fitting. These methods are inefficient and ignore global context when learning to segment lanes~\cite{PINet,SCNN}. To tackle the context learning issue, some methods~\cite{SCNN,LaneAndRoad,LaneAndGan} use message passing or extra scene annotations to capture the global context for enhancement of final performance, but these methods inevitably consume more time and data cost~\cite{ENet-SAD}. Unlike these methods, a soft attention based method~\cite{ENet-SAD} generates a spatial weighting map that distills a richer context without external consumes. However, the weighting map only measures the feature's importance, limiting its usage to consider the dependencies between features that support to infer slender structures. On the other hand, to improve the algorithm's efficiency, 
\cite{Line-CNN} transfers the pipeline in object detection to detect lanes without the above segmentation procedure and post-processing, but it relies on complex anchor designing choices and additional non-maximal suppression, making it even slower than most lane detectors. Recently, a method~\cite{PolyLaneNet} reframes the task as lane markings fitting by polynomial regression, which achieves significant efficiency but still has a large accuracy gap with other methods due to the neglect of learning the global context.

To tackle these issues, we propose to reframe the lane detection output as parameters of a lane shape model and develop a network built with non-local building blocks to reinforce the learning of global context and lanes' slender structures.
The output for each lane is a group of parameters which approximates the lane marking with an explicit mathematical formula derived from road structures and the camera pose. 
Given specific priors such as camera intrinsic, those parameters can be used to calculate the road curvature and camera pitch angle without any 3D sensors.
Next, inspired by natural language processing models which widely employ transformer block~\cite{AttentionIsAllYourNeed} to explicitly model long-range dependencies in language sequence,
we develop a transformer-based network that summarizes information from any pairwise visual features, enabling it to capture lanes' long and thin structures and global context.
The whole architecture predicts the proposed outputs at once and is trained end-to-end with a Hungarian loss. The loss applies bipartite matching between predictions and ground truths to ensure one-to-one disorderly assignment, enabling the model to eliminate an explicit non-maximal suppression process.

The effectiveness of the proposed method is validated in the conventional TuSimple lane detection benchmark~\cite{WebTuSimple}. Without bells and whistles, our method achieves state-of-the-art accuracy and the lowest false positive rate with the most lightweight model size and fastest speed.
In addition, to evaluate the adaptability to new scenes, we collect a large scale challenging dataset called Forward View Lane (FVL) in multiple cities across various scenes (urban and highway, day and night, various traffic and weather conditions). Our method shows strong adaptability to new scenes even that the TuSimple dataset does not contain, e.g., night scenes.

The main contributions of this paper can be summarized as follows:
\begin{itemize}
    \item We propose a lane shape model whose parameters serve as directly regressed output and reflect road structures and the camera pose.
    \item We develop a transformer-based network that considers non-local interactions to capture long and thin structures for lanes and the global context.
    \item Our method achieves state-of-the-art accuracy with the least resource consumption and shows excellent adaptability to a new challenging self-collected lane detection dataset.
\end{itemize}

\section{Related Work}
The authors of~\cite{LDWReview} provide a good overview of the techniques used in traditional lane detection methods. Feature-based methods usually extract low-level features (lane segments) by Hough transform variations, then use clustering algorithms such as DBSCAN (Density Based Spatial Clustering of Applications with Noise) to generate final lane detections~\cite{LaneFeatureMethod}. Model-based methods use top-down priors such as geometry and road surface~\cite{LaneModelMethod}, which describe lanes in more detail and shows excellent simplicity. 

In recent years, methods based on deep neural networks have been shown to outperform traditional approaches.
Earlier methods~\cite{LaneNet,FastDraw} typically extract dense segmentation results and then employ post-processing such as segment clustering and curve fitting. Their performances are limited by the initial segmentation of lanes due to the difficulties of learning such long and thin structures. 
To address this issue, SCNN~\cite{SCNN} uses message passing to capture a global context, exploiting richer spatial information to infer occluded parts.
~\cite{LaneAndRoad,LaneAndGan} adopt extra scene annotations to guide the model's training, which enhances the final performance.
Unlike them that need additional data and time cost, ENet-SAD~\cite{ENet-SAD} applies a soft attention mechanism, which generates weighting maps to filter out unimportant features and distill richer global information.
In contrast to inferring based on dense segmentation results, PINet~\cite{PINet} extracts a sparse point cloud to save the computations, but it also requires inefficient post-processing like outlier removal.

In contrast to these methods, our method directly outputs parameters of a lane shape model. The whole method works in an end-to-end fashion without any intermediate representation or post-processing.

In literature, some end-to-end lane detectors have been proposed recently.
Line-CNN~\cite{Line-CNN} transfers the success of Faster-RCNN~\cite{FasterRCNN} into lane detection by predicting offsets based on pre-designed straight rays (like anchor boxes), which achieves state-of-the-art accuracy. However, it inevitably suffers the drawbacks of complex ad-hoc heuristics choices to design rays and additional non-maximal suppress, making it even slower than most lane detectors.
PolyLaneNet~\cite{PolyLaneNet} reframes the lane detection task as a polynomial regression problem, which achieved the highest efficiency. However, its accuracy still has a large gap with other methods, especially has difficulty in predicting lane lines with accentuated curvatures due to neglect of learning global information and ignorant of road structures modeling.

In this work, our approach also expects a parametric output but differs in that these parameters are derived from a lane shape model which models the road structures and the camera pose. These output parameters have explicit physical meanings rather than simple polynomial coefficients. In addition, our network is built with transformer block~\cite{AttentionIsAllYourNeed} that performs attention in modeling non-local interactions, enabling it to reinforce the capture of long and thin structures for lanes and learning of global context information.

\section{Method}
Our end-to-end method reframes the output as parameters of a lane shape model. Parameters are predicted by using a transformer-based network trained with a Hungarian fitting loss. 

\subsection{Lane Shape Model}
The prior model of the lane shape is defined as a polynomial on the road. Typically, a cubic curve is used to approximate a single lane line on flat ground:
\begin{equation}
\label{poly3order}
X=kZ^3+mZ^2+nZ+b,
\end{equation}
where $k, m, n$ and $b$ are real number parameters, $k \neq 0$. The $\left(X, Z\right)$ indicates the point on the ground plane. When the optical axis is parallel to the ground plane, the curve projected from the road onto the image plane is:
\begin{equation}
\label{imagecurve}
u = \frac{k'}{v^2} + \frac{m'}{v} + n' + b' \times v,
\end{equation}
where $k',m',n',b'$ are composites of parameters and camera intrinsic and extrinsic parameters, and $\left(u,v\right)$ is a pixel at the image plane.

In the case of a tilted camera whose optical axis is at an angle of $\phi$ to the ground plane, the curve transformed from the untitled image plane to the tilted image plane is:
\begin{equation}
\label{tiltedcurve}
\begin{split}
u' = & \frac{k' \times \cos^2 \phi}{\left(v' - f \sin \phi\right)^2} + \frac{m' \cos \phi}{\left(v' - f \sin \phi\right)} + n' \\
+ & \frac{b' \times v'}{\cos \phi} - b' \times f \tan \phi,
\end{split}
\end{equation}
here $f$ is the focal length in pixels, and $\left(u',v'\right)$ is the corresponding pitch-transformed position. When $\phi=0$, the curve function Eq.~\ref{tiltedcurve} will be simplified to Eq.~\ref{imagecurve}. Details of the derivation can be reviewed in Sec.~\ref{appendix}.

\noindent \textbf{Curve re-parameterization.}
By combining parameters with the pitch angle $\phi$, the curve in a tilted camera plane has the form of:
\begin{equation}
\label{combtiltedcurve}
u' = \frac{k''}{\left(v'-f''\right)^2} + \frac{m''}{\left(v' - f''\right)} + n' + b'' \times v' - b''',
\end{equation}
here, the two constant terms $n'$ and $b'''$ are not integrated because they contain different physical parameters.

Apart from that, the vertical starting and ending offset $\alpha, \beta$ are also introduced to parameterize each lane line. These two parameters provide essential localization information to describe the upper and lower boundaries of lane lines.

In real road conditions, lanes typically have a global consistent shape. Thus, the approximated arcs have a equal curvature from the left to the right lanes, so $k'', f'', m'', n'$ will be shared for all lanes.
Therefore, the output for the $t$-th lane is re-parameterized to $g_t$:
\begin{equation}
\label{shapemodel}
g_t=\left(k'', f'', m'', n', b''_{t}, b'''_{t}, \alpha_t, \beta_t\right)
\end{equation}
where $t\in\left\{1, ..., T\right\}$, $T$ is the number of lanes in an image. Each lane only differs in bias terms and lower/upper boundaries. 

\begin{figure*}[t]
\begin{center}
\includegraphics[width=174mm]{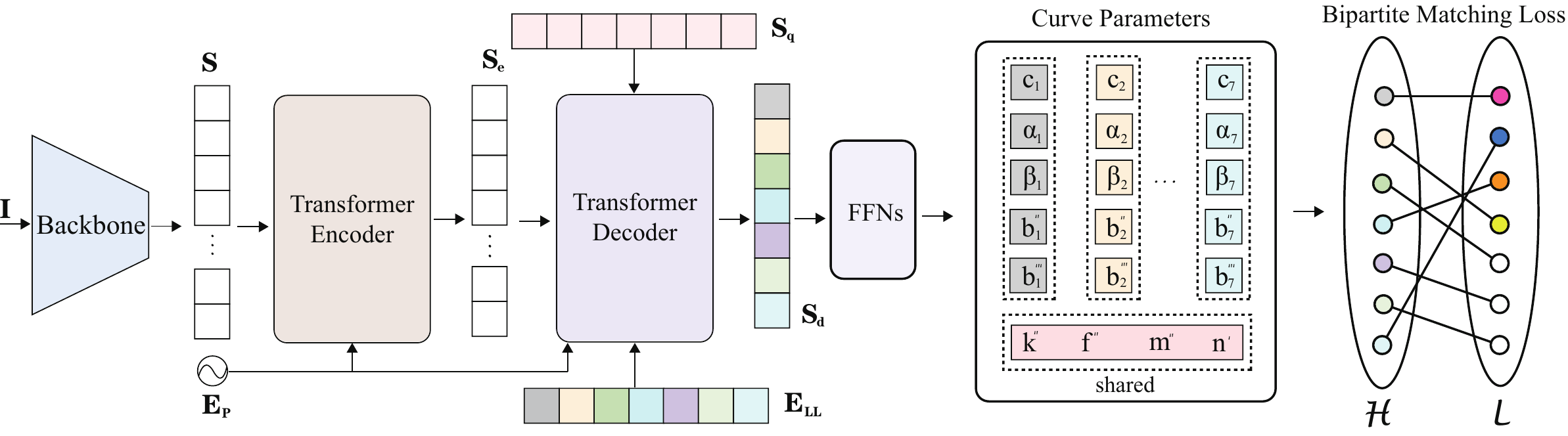}
\end{center}
\caption{{Overall Architecture.} The $\mathbf{S}$, $\mathbf{S_e}$ and $\mathbf{E_p}$ indicate flattened feature sequence, encoded sequence and the sinusoidal positional embeddings which are all tensors with shape $HW\times C$.
The $\mathbf{S}_q$, $\mathbf{E_{LL}}$ and $\mathbf{S_d}$ represent query sequence, learned lane embedding and the decoded sequence which are all in shape $N \times C$. Different color indicate different output slots. White hollow circles represent "non-lanes".}
\label{fig:modelandloss}
\end{figure*}

\begin{figure}[t]
\begin{center}
\includegraphics[width=83mm]{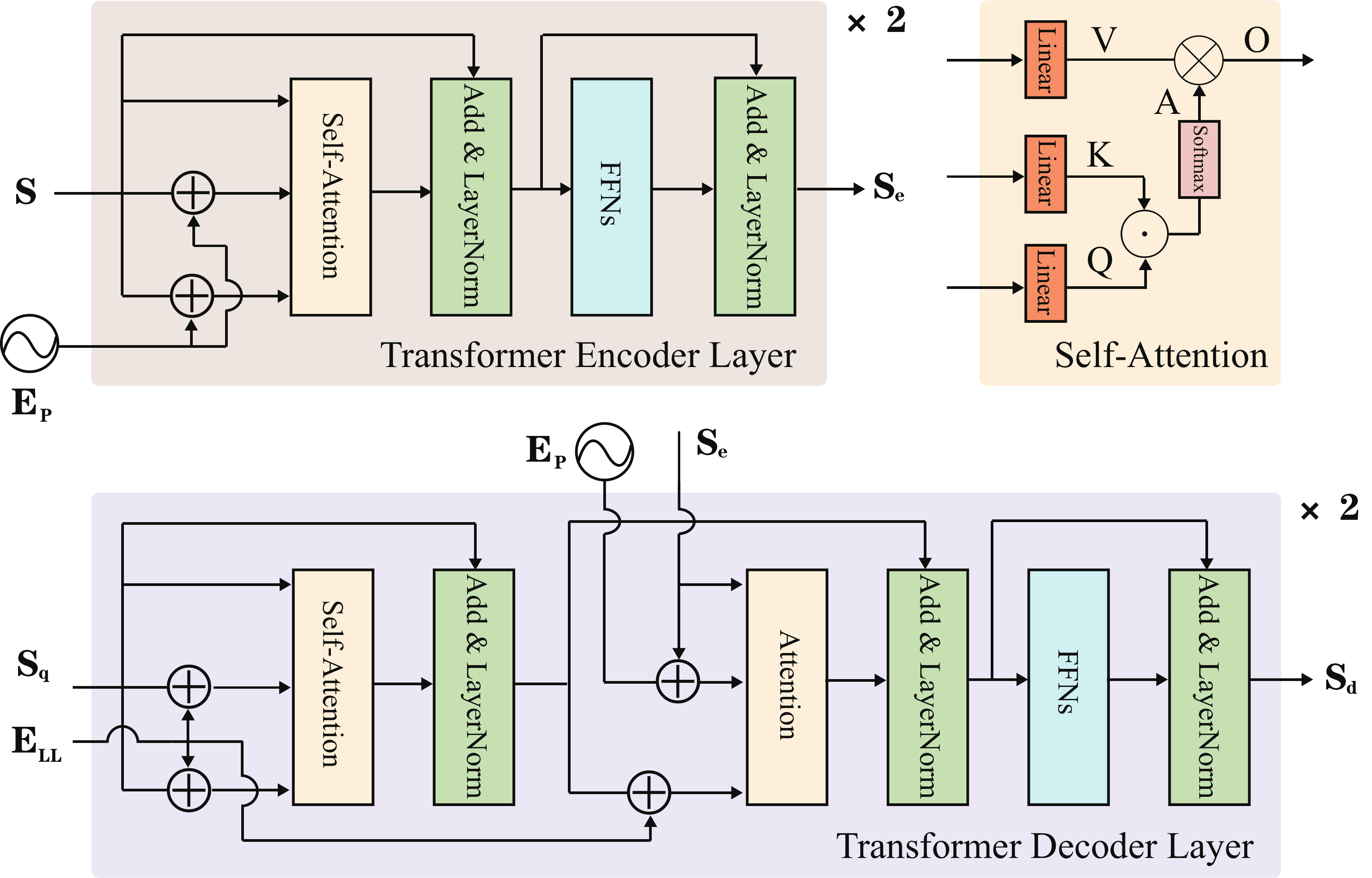}
\end{center}
\caption{{Transformer Encoder and Decoder.} The $\oplus$ and $\odot$ represent matrix addition and dot-product operations respectively.}
\label{fig:encdec}
\end{figure}

\subsection{Hungarian Fitting Loss}
The Hungarian fitting loss performs a bipartite matching between predicted parameters and ground truth lanes to find positives and negatives. The matching problem is efficiently solved by the Hungarian algorithm. Then the matching result is used to optimize lane-specific regression losses.

\noindent \textbf{Bipartite matching.}
Our method predicts a fixed $N$ curves, where $N$ is set to be larger than the maximum number of lanes in the image of a typical dataset.
Let us denote the predicted curves by $\mathcal{H}=\left\{h_i|h_i=\left(c_i, g_i\right)\right\}_{i=1}^{N}$, where $c_i \in \left\{0, 1\right\}$ (0: non-lane, 1: lane). The ground truth lane marking is represented by a sequence $\mathbf{\hat{s}}=\left(\hat{u}'_{r}, \hat{v}'_{r}\right)_{r=1}^{R}$, where $r$ sequentially indexes the sample point within range $R$ and $\hat{v}'_{r+1} > \hat{v}'_{r}$. Since the number of predicted curves $N$ is larger than the number of ground truth lanes, we consider the ground truth lanes also as a set of size $N$ padded with non-lanes $\mathcal{L}=\left\{\hat{l}_i|\hat{l}_i=\left(\hat{c}_i, \hat{\mathbf{s}}_i\right)\right\}_{i=1}^{N}$.
We formulate the bipartite matching between the set of curves and the set of ground truth lane markings as a cost minimization problem by searching an optimal injective function $z: \mathcal{L} \rightarrow \mathcal{H}$, i.e., $z\left(i\right)$ is the index of curve assigned to fitting ground-truth lane $i$:
\begin{equation}
\label{optimalproblem}
\hat{z}=\mathop{\arg\min}_{z} \sum_{i=1}^N d\left(\hat{l}_i, h_{z\left(i\right)}\right),
\end{equation}
where $d$ measures the matching cost given a specific permutation $z$ between the $i$-th ground truth lane and a predicted curve with index $z\left(i\right)$. Following prior work (e.g., ~\cite{PDCrowdedScenes}), this problem can be solved efficiently by the Hungarian algorithm.

For the prediction with index $z\left(i\right)$, the probability of class $\hat{c}_i$ is defined as $p_{z\left(i\right)}\left(\hat{c}_i\right)$, and the fitting lane sequence is defined as $\mathbf{s}_{z\left(i\right)}=\left(u'_{ri}, \hat{v}'_{ri}\right)_{r=1}^{R_i}$, where $R_i$ is the length of $i$-th lane and $u'_{ri}$ is calculated using Eq.~\ref{combtiltedcurve} based on the predicted group of parameters $g_i$. Then the matching cost $d$ has the form of:
\begin{equation}
\label{matchingcost}
\begin{split}
d = & - \omega_1 p_{z\left(i\right)}\left(\hat{c}_i\right) + \mathbbm{1}\left(\hat{c}_i = 1\right) \omega_2 L_{1} \left(\hat{\mathbf{s}}_i, \mathbf{s}_{z\left(i\right)} \right) \\
& + \mathbbm{1}\left(\hat{c}_i = 1\right) \omega_3 L_{1} \left(\hat{\alpha}_i, \mathbf{\alpha}_{z\left(i\right)}, \hat{\beta}_i, \mathbf{\beta}_{z\left(i\right)} \right),
\end{split}
\end{equation}
where $\mathbbm{1}\left(\cdot\right)$ is an indicator function, $\omega_1$, $\omega_2$ and $\omega_3$ adjusts the effect of the matching terms, and $L_{1}$ is the commonly-used mean absolute error. We use the probabilities instead of log-probabilities following~\cite{DETR} because this makes the classification term commensurable to the curve fitting term.

\noindent \textbf{Regression loss.}
The regression loss calculates the error for all pairs matched in the previous step with the form of:
\begin{equation}
\label{realloss}
\begin{split}
L = \sum_{i=1}^N & -\omega_1 \log p_{\hat{z}\left(i\right)}\left(\hat{c}_i\right) + \mathbbm{1}\left(\hat{c}_i = 1\right) \omega_2 L_{1} \left(\hat{\mathbf{s}}_i, \mathbf{s}_{\hat{z}\left(i\right)} \right) \\
& + \mathbbm{1}\left(\hat{c}_i = 1\right) \omega_3 L_{1} \left(\hat{\alpha}_i, \mathbf{\alpha}_{\hat{z}\left(i\right)}, \hat{\beta}_i, \mathbf{\beta}_{\hat{z}\left(i\right)} \right),
\end{split}
\end{equation}
where $\hat{z}$ is the optimal permutation calculated in Eq.~\ref{matchingcost}. The $\omega_1, \omega_2$, and $\omega_3$ also adjust the effect of the loss terms and are set to be the same values of coefficients in Eq.~\ref{matchingcost}.

\subsection{Architecture}
The architecture shown in Fig.~\ref{fig:modelandloss} consists of a backbone, a reduced transformer network, several feed-forward networks (FFNs) for parameter predictions, and the Hungarian Loss. Given an input image $\mathbf{I}$, the backbone extracts a low-resolution feature then flattens it into a sequence $\mathbf{S}$ by collapsing the spatial dimensions. 
The $\mathbf{S}$ and positional embedding $\mathbf{E}_{\mathbf{p}}$ are fed into the transformer encoder to output a representation sequence $\mathbf{S_\mathbf{e}}$.
Next, the decoder generates an output sequence $\mathbf{S}_{\mathbf{d}}$ by first attending to an initial query sequence $\mathbf{S}_{\mathbf{q}}$ and a learned positional embedding $\mathbf{E_\mathbf{LL}}$ that implicitly learns the positional differences, then computing interactions with $\mathbf{S_\mathbf{e}}$ and $\mathbf{E}_{\mathbf{p}}$ to attend to related features.
Finally, several FFNs directly predict the parameters of proposed outputs.

\noindent \textbf{Backbone.} The backbone is built based on a reduced ResNet18. The original ResNet18~\cite{ResNet} has four blocks and downsamples features by 16 times. The output channel of each block is "64, 128, 256, 512". Here, our reduced ResNet18 cuts the output channels into "16, 32, 64, 128" to avoid overfitting and sets the downsampling factor as 8 to reduce losses of lane structural details. Using an input image $\mathbf{I}$ as input, the backbone extracts a low-resolution feature that encodes high-level spatial representations for lanes with a size of $H \times W \times C$. Next, to construct a sequence as the input of encoder, we flatten that feature in spatial dimensions, resulting in a sequence $\mathbf{S}$ with the size of $ HW \times C$, where $HW$ denotes the length of the sequence and $C$ is the number of channels.

\noindent \textbf{Encoder.} The encoder has two standard layers that are linked sequentially. Each of them consists of a self-attention module and a feed-forward layer shown in~Fig.\ref{fig:encdec}. Given the sequence $\mathbf{S}$ that abstracts spatial representations, the sinusoidal embeddings $\mathbf{E}_\mathbf{p}$ based on the absolute positions~\cite{AttentionIsAllYourNeed} is used to encode positional information to avoid the permutation equivariant. The $\mathbf{E}_\mathbf{p}$ has the same size as $\mathbf{S}$. The encoder performs scaled dot-product attention by Eq.~\ref{attention}:
\begin{equation}
\label{attention}
\mathbf{A}=\mbox{softmax}\left(\frac{\mathbf{Q}\mathbf{K}^{\rm T}}{\sqrt{C}}\right),~\mathbf{O}=\mathbf{A}\mathbf{V},
\end{equation}
where the $\mathbf{Q},\mathbf{K},\mathbf{V}$ denote sequences of query, key and value through a linear transformation on each input row, and $\mathbf{A}$ represents the attention map which measures non-local interactions to capture slender structures plus global context, and $\mathbf{O}$ indicates the output of self-attention. The output sequence of encoder $\mathbf{S}_\mathbf{e}$ with the shape of $HW \times C$ is obtained by following FFNs, residual connections with layer normalizations~\cite{layernorm}, and another same encoder layer.

\noindent \textbf{Decoder.} The decoder also has two standard layers. Unlike the encoder, each layer inserts the other attention module, which expects the output of the encoder, enabling it to perform attention over the features containing spatial information to associate with the most related feature elements. Facing the translation task, the original transformer~\cite{AttentionIsAllYourNeed} shifts the ground truth sequence one position as the input of the decoder, making it output each element of the sequence in parallel at a time. In our task, we just set the input $\mathbf{S}_{\mathbf{q}}$ as an empty $N \times C$ matrix and directly decodes all curve parameters at a time. Additionally, we introduce a learned lane embedding $\mathbf{E}_\mathbf{LL}$ with the size of $N \times C$, which serves as a positional embedding to implicitly learn global lane information. The attention mechanism works with the same formula~Eq.~\ref{attention} and the decoded sequence $\mathbf{S}_{\mathbf{d}}$ with the shape of $N \times C$ is obtained sequentially like the way in the encoder. When training, intermediate supervision is applied after each decoding layer.

\noindent \textbf{FFNs for Predicting Curve Parameters.} The prediction module generates the set of predicted curves $\mathcal{H}$ using three parts. A single linear operation directly projects the $\mathbf{S}_{\mathbf{d}}$ into $N \times 2$ then a softmax layer operates it in the last dimension to get the predicted label (background or lane) $c_{i}, i \in \left\{1,\dots,N\right\}$. Meanwhile, one 3-layer perceptron with ReLU activation and hidden dimension $C$ projects the $\mathbf{S}_{\mathbf{d}}$ into $N \times 4$, where dimension $4$ represents four groups of lane-specific parameters. The other 3-layer perceptron firstly projects a feature into $N \times 4$ then averages in the first dimension, resulting in the four shared parameters.

\section{Experiments}
\begin{figure*}[t]
\begin{center}
\includegraphics[width=174mm]{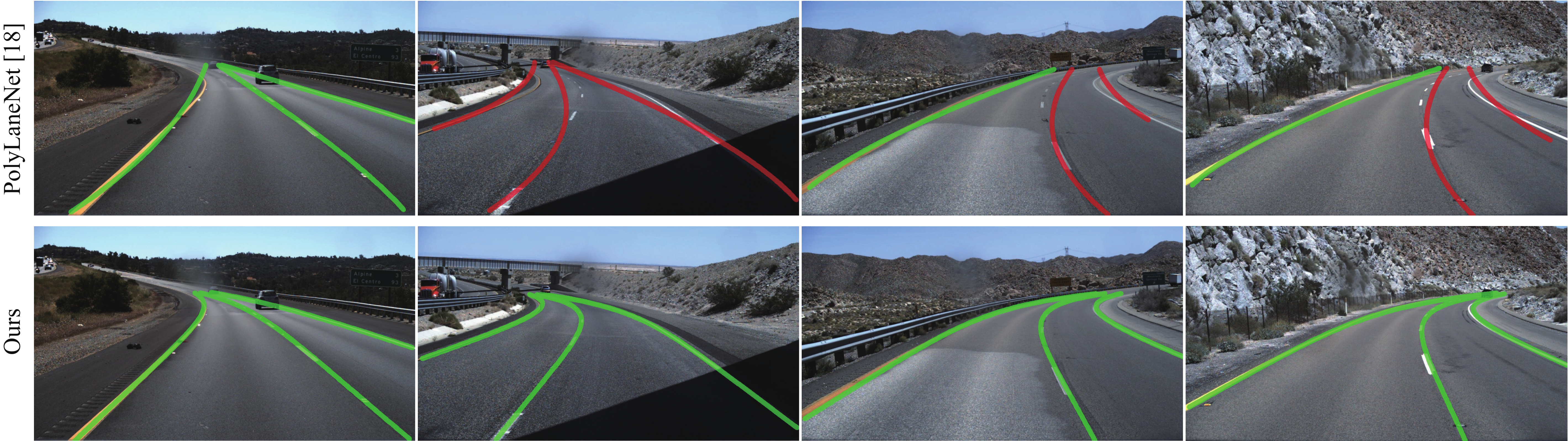}
\end{center}
\caption{{Qualitative comparative results on TuSimple test set.} The first row visualizes the predicted curves by the best model of officially public PolyLaneNet resources (red curves means these predictions are mismatched). The second row visualizes our predictions.}
\label{fig:comparepolylanenet}
\end{figure*}

\noindent \textbf{Datasets.} The widely-used TuSimple~\cite{WebTuSimple} lane detection dataset is used to evaluate our method. The TuSimple dataset consists of 6408 annotated images which are the last frames of video clips recorded by a high-resolution ($720 \times 1280$) forward view camera across various traffic and weather conditions on America's highways in the daytime. It is split initially into a training set (3268), a validation set (358), and a testing set (2782). 
To evaluate the adaptive capability to new scenes, we introduce a much more complex self-collected dataset named Forward View Lane (FVL). The FVL contains 52970 images with a raw resolution of $720 \times 1280$. These images were collected by a monocular forward-facing camera typically located near the rear-view mirror in multiple cities across different scenes (urban and highway, day and night, various traffic and weather conditions). The FVL contains more challenging road conditions and will go public to help research for the community.

\noindent \textbf{Evaluation Metrics.}
To compare the performance against previous methods, we follow the literature and calculate the accuracy using TuSimple metrics. The prediction accuracy is computed as $Accuracy=\frac{\sum_{vc} TPr_{vc}}{\sum_{vc} Gt_{vc}}$, where $TPr_{vc}$ is the number of true prediction points in the last frame of the video clip, and $Gt_{vc}$ is the number of ground truth points. A point is considered as a true positive if its distance from the corresponding label point is within 20 pixels as the TuSimple benchmark suggested~\cite{PolyLaneNet}. Besides, false positives (FP) and false negatives (FN) rates are also reported~\cite{WebTuSimple}.

\noindent \textbf{Implementation Details.} The hyperparameter settings are the same for all experiments except for the ablation study. The input resolution is set to $360 \times 640$. The raw data are augmented by random scaling, cropping, rotating, color jittering, and horizontal flipping. The learning rate is set to be 0.0001 and decayed 10 times every 450k iterations. Batch size is set as 16, and loss coefficients $\omega_1$, $\omega_2$ and $\omega_3$ are set as 3, 5 and 2. The fixed number of predicted curves $N$ is set as 7, and the number of training iterations is set as 500k. All those hyper-parameters are determined by maximizing the performance on the TuSimple validation set.

In the following section, we treat PolyLaneNet~\cite{PolyLaneNet} as the baseline method since they also predict parametric output for lanes and provide amazingly reproducible codes and baseline models. Besides, to best show our performance, we also compare with other state-of-the-art methods PINet~\cite{PINet}, Line-CNN~\cite{Line-CNN}, ENet-SAD~\cite{ENet-SAD}, SCNN~\cite{SCNN}, FastDraw~\cite{FastDraw}. The proposed method was trained using both TuSimple training and validation set as previous works did. The time unit compares the FPS performance, and we also report MACs and the total number of parameters. All results are tested on a single GTX 1080Ti platform.

\subsection{Comparisons with State-of-the-Art Methods}
\begin{table}[t]
\begin{center}
\caption{Comparisons of accuracy~(\%) on TuSimple testing Set. The number of multiply-accumulate (MAC) operations is given in G. The number of parameters (Para) is given in M (million). The PP means the requirement of post-processing.}
\setlength{\tabcolsep}{0.3mm}{
\begin{tabular}{|l|c|c|c|c|c|c|c|}
\hline
Method      & FPS  & MACs  & Para & PP & Acc & FP & FN \\
\hline\hline
FastDraw~\cite{FastDraw}    & 90   & -     & -       & \checkmark & 95.20 & .0760 & .0450    \\
SCNN~\cite{SCNN}        & 7    & -     & 20.72 & \checkmark & 96.53 & .0617 & \textbf{.0180}   \\
ENet-SAD~\cite{ENet-SAD}    & 75   & -     & 0.98  & \checkmark & 96.64 & .0602 & .0205      \\
PINet~\cite{PINet}       & 30   & -     & 4.39  & \checkmark & 96.70 & \textbf{.0294} & .0263   \\
Line-CNN~\cite{Line-CNN}    & 30   & -     & -       & -     & \textbf{96.87} & .0442 & .0197 \\
\hline\hline
PolyLaneNet~\cite{PolyLaneNet} & 115  & 1.784 & 4.05  & -    & 93.36 & .0942  & .0933     \\
Ours        & \textbf{420} & \textbf{0.574}  & \textbf{0.77} & -    & \textbf{96.18} & \textbf{.0291} & \textbf{.0338} \\
\hline
\end{tabular}}
\label{tab:TuSimpleResult}
\end{center}
\end{table}

Tab.~\ref{tab:TuSimpleResult} shows the performance on TuSimple benchmark. Without bells and whistles, our methods outperforms the PolyLaneNet by \textbf{2.82\%} accuracy with \textbf{5} $\mathbf{\times}$ fewer parameters and runs \textbf{3.6} $\mathbf{\times}$ faster. 
Compared to the state-of-the-art method Line-CNN, ours is only 0.69\% lower in accuracy but runs \textbf{14} $\mathbf{\times}$ faster than it.
Compared with other two stages approaches, our method achieves competitive accuracy and the lowest false positive rate with the fewest parameters and much faster speed. The high false positive rate would lead to more severe risks like false alarming and rapid changes than missing detections in real applications~\cite{PINet}. In a word, our method has tremendous mobile deployment capabilities than any other algorithms.

The visualization of the lane detection results is given in Fig.~\ref{fig:comparepolylanenet}. By comparing the areas closer to the horizon, our method is capable of catching structures with fewer details, showing an excellent performance on lane markings with large amplitude curvature than the baseline method. We attribute this to (1) the global lane shape consistency implicitly requires the model to learn consistent shapes across all supervision information from all lane markings; (2) the attention mechanism does capture non-local information, supplementing the contextual information for the missing details, which helps capture slender structures. Subsequent ablation experiments further corroborate these conclusions.

\begin{figure}[t]
\begin{center}
\includegraphics[width=83mm]{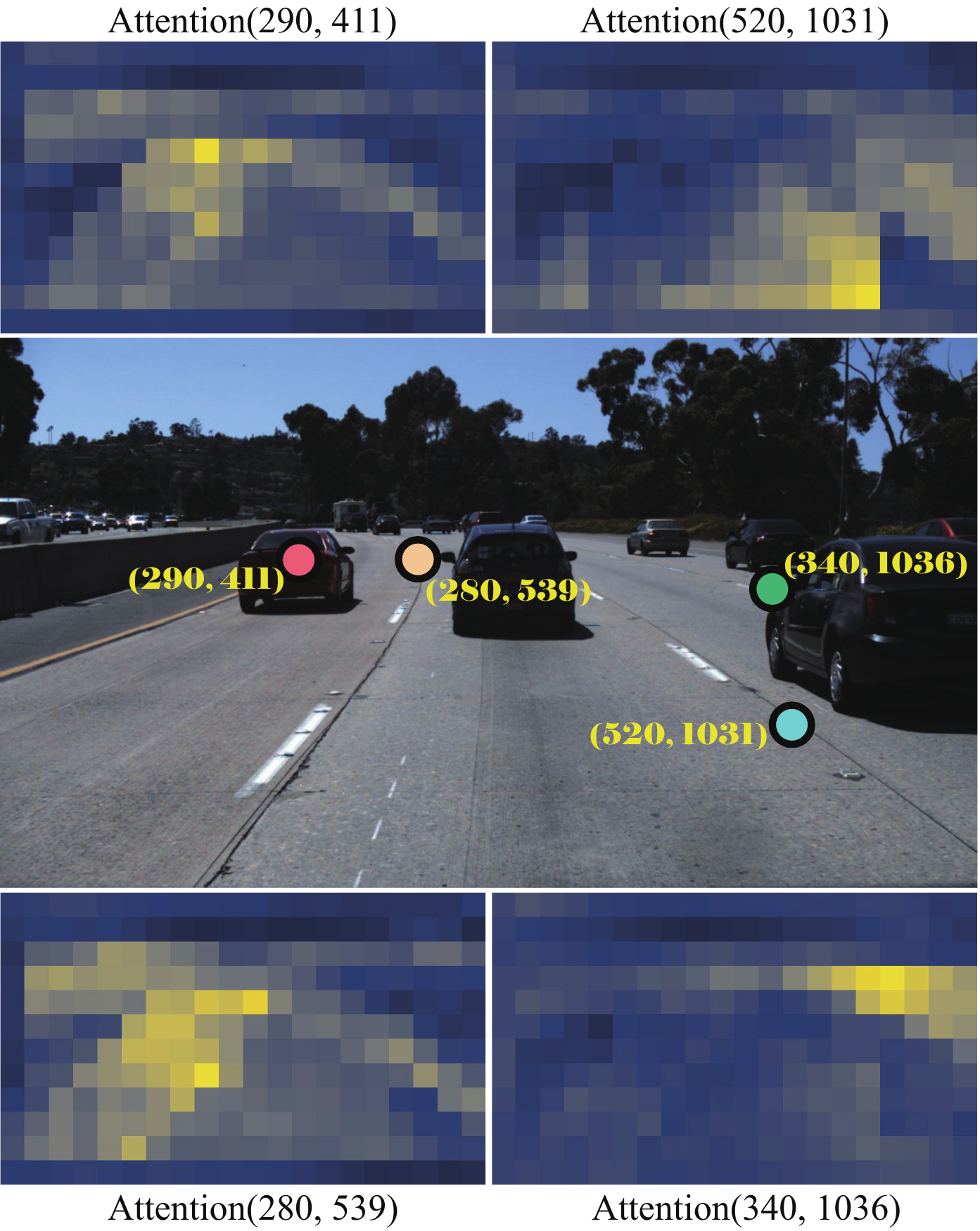}
\end{center}
\caption{{Encoder attention maps for different sampling points.} The encoder seems to aggregate a lot contextual information and capture slender structures.}
\label{fig:encattnmap}
\end{figure}

\begin{figure}[t]
\begin{center}
\includegraphics[width=83mm]{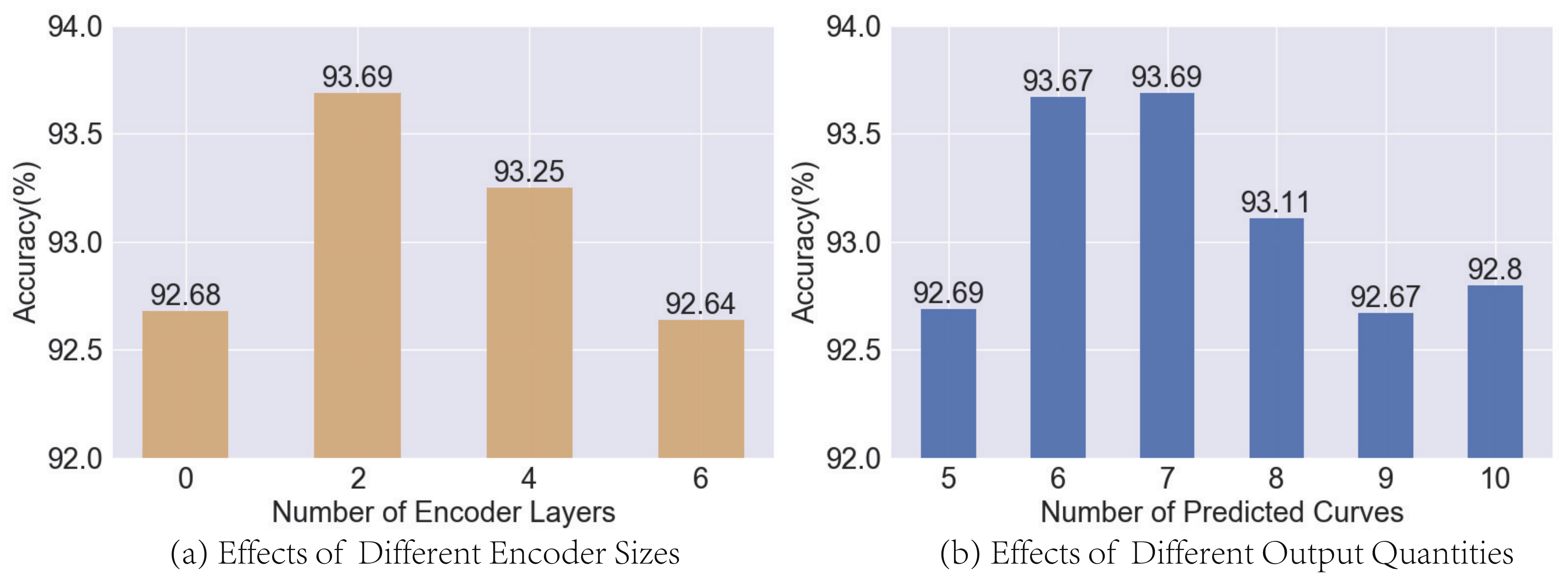}
\end{center}
\caption{{Quantitative evaluation of (a) encoders size and  (b) number of predictions on TuSimple validation set~(\%).} The decoder size is fixed to 2.}
\label{fig:encAndEllsize}
\end{figure}

\begin{figure}[t]
\begin{center}
\includegraphics[width=83.5mm]{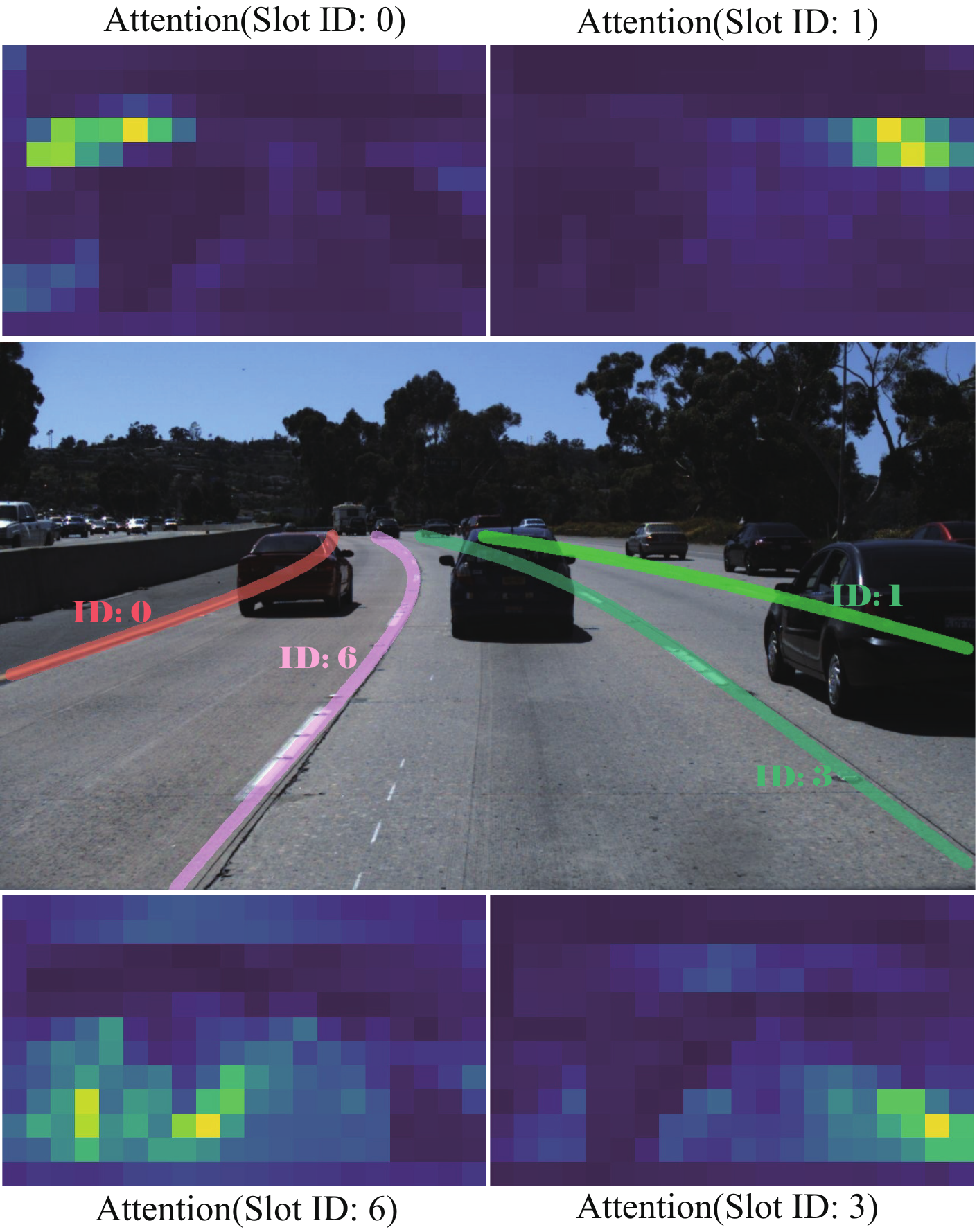}
\end{center}
\caption{{Decoder attention maps for output slots.} The decoder mainly focus on local structures.}
\label{fig:decattnmap}
\end{figure}

\begin{figure*}[t]
\begin{center}
\includegraphics[width=174mm]{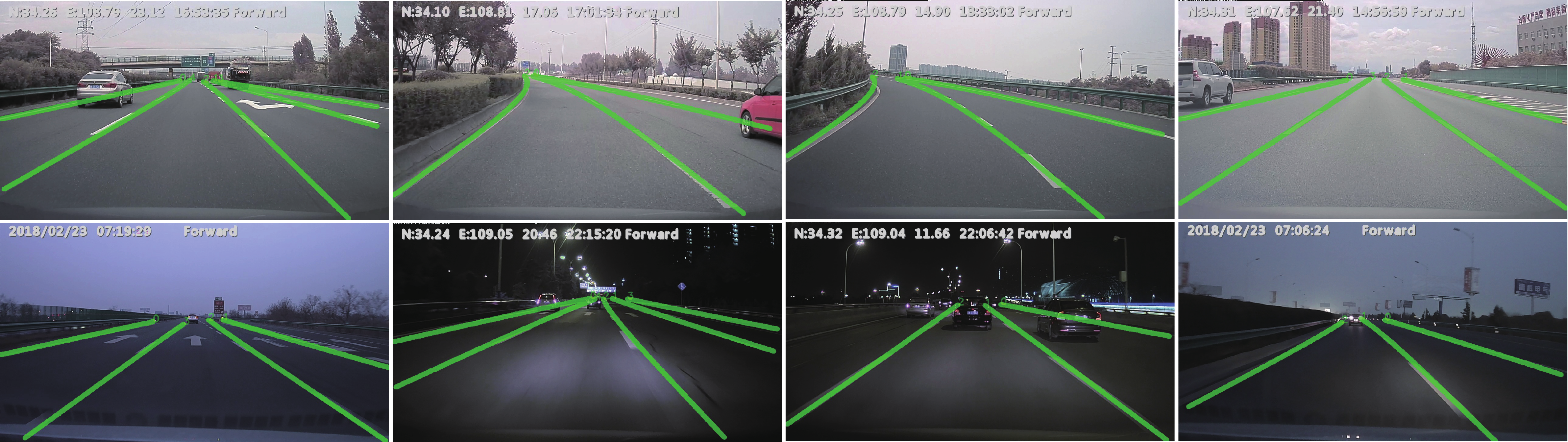}
\end{center}
\caption{{Qualitative transfer results on FVL dataset.} Our method even estimates exquisite lane lines without ever seeing the night scene.}
\label{fig:fvvviz}
\end{figure*}

\subsection{Ablation Study}
\noindent \textbf{Investigation of Shape Model.}
\begin{table}[t]
\begin{center}
\caption{Quantitative evaluation of different shape models on TuSimple validation set~(\%).}
\setlength{\tabcolsep}{1.9mm}{
\begin{tabular}{|l|c|c|c|c|}
\hline
Curve Shape    & Consistency & Acc   & FP & FN         \\
\hline\hline
Quadratic      & -           & 91.94 & 0.1169 & 0.0975 \\
Quadratic      & \checkmark  & 93.18 & 0.1046 & 0.0752 \\
Cubic          & -           & 92.64 & 0.1068 & 0.0868 \\
Cubic          & \checkmark  & \textbf{93.69} & \textbf{0.0979} & \textbf{0.0724} \\
\hline
\end{tabular}}
\label{tab:globalorlocal}
\end{center}
\end{table}
To investigate the effect of the lane shape model, we tested different shape models. The comparison results are listed in Tab.~\ref{tab:globalorlocal}. The header 'Curve Shape' denotes our hypothetical approximation of the lanes on the roadway. Without shape consistency constraint, the $i$-th predicted curve has its own predicted shape parameters, $k''_{i}, f''_{i}, m''_{i}, n'_{i}$, regressed by a 3-layer perceptron without averaging.
$i\in\left\{1, ..., N\right\}$, $N$ is the number of predictions.

From Tab.~\ref{tab:globalorlocal}, we find that the best model is using cubic curve approximation with the shape consistency. The consensus in the lane detection field is that high order lane models always fit lane markings better than simple models~\cite{LaneModeling,CubicCurve}. Besides, the shape consistency further improves accuracy. We attribute this to the effect of sample points equilibrium. In the TuSimple dataset, a single lane line close to the horizon is marked using far fewer points than the one close to the camera, and this imbalance makes the model more biased towards a better performance at the nearby areas~\cite{PolyLaneNet}. To tackle this issue, the shape consistency requires the model to fit the same curvature at areas close to the horizon, enabling it to use all remote points to infer the same curvature which is appropriate for all lanes.

\noindent \textbf{Number of encoder layers.}
To investigate the effects of performing attention on spatial features, we tested different numbers of encoder layers. From Fig.~\ref{fig:encAndEllsize}(a), without self-attention mechanism, the accuracy drops by 1.01\%. Meanwhile, we found that the attention mechanism could be overused. When the same accuracy of the training set ( $\approx$ 96.2\%) was observed, a larger number led to a degradation of the model generalization performance. It appears that our model is approaching the capacity limit of the data's expressive ability.

To better understand the effect of the encoder, we visualize the attention maps $\mathbf{A}$ of the last encoder layer in Fig.~\ref{fig:encattnmap}. We tested a few points on the lane markings with different conditions.
The orange-red lane marking point is totally occluded, so the encoder seems to focus on its right unobstructed lane line features.
The peach-puff point is located with clear lane markings, and its attention map shows a clear long and thin structure for the lane.
Despite the lack of appearance clues in the local neighborhood of the light-blue point, the encoder still identifies a distinct slender structure by learning the global context (distant markings and nearby vehicles).
Similarly, the green point misses many details, but the encoder still identifies a relevant long and thin structure by learning the global context.

\noindent \textbf{Number of decoder layers.}
\begin{table}[t]
\begin{center}
\caption{Quantitative evaluation of decoder size and different decoder layer on TuSimple validation set~(\%). The encoder size is set to be 2.}
\setlength{\tabcolsep}{1.0mm}{
\begin{tabular}{|l|c|c|c|c|c|c|}
\hline
\diagbox{Size}{Layer} & 1     & 2     & 3 & 4 & 5 & 6    \\
\hline\hline
2            & 93.55 & \textbf{93.69} & - & - & - & - \\
4            & 92.52 & 93.08 & 93.15 & \textbf{93.15} & - & - \\
6            & 92.70 & 93.07 & 93.05 & 93.13 & 93.14 & \textbf{93.16} \\
\hline
\end{tabular}}
\label{tab:decodersize}
\end{center}
\end{table}
To investigate the performance of auxiliary losses, we changed the number of decoder layers. From Tab.~\ref{tab:decodersize}, the output of the last layer is the highest in each configuration, while as the layers become more numerous, the overall performance gradually degrades due to overfitting. 

Similarly to visualizing encoder attention, we analyze the attention map for each output of the decoder in Fig.~\ref{fig:decattnmap}. We observe that decoder attention mainly focuses on its own slender structures which help the model to separate specific lane instances directly rather than using additional non-maximal suppression.

\noindent \textbf{Number of predicted curves.}
The predicted curves play a similar role as the anchor boxes~\cite{FasterRCNN}, which generate positive and negative samples based on some matching rules. The difference is that we only find one-to-one matching without duplicates, so the number of predicted curves determines the number of negative samples. To investigate the impact of positive and negative sample proportions, we tested from 5 (the TuSimple has up to 5 ground truth lane markings in one image) to 10 in increments of 1.

From Fig.~\ref{fig:encAndEllsize}(b), the best size is 7. As the number of predicted curves gets smaller, the network gradually loses its generalization capability because the lack of negatives makes the training inefficient, resulting in degenerate models~\cite{FocalLoss}. Moreover, as the number increases, the performance also degrades since lane curve fitting needs a sufficient number of positives. More negatives wrongly guide the model to optimizing loss by paying more attention to the classification for negatives, which weakens the performance of curve fitting for positives.

\subsection{Transfer Results on FVL Dataset}
Fig.~\ref{fig:fvvviz} demonstrates the qualitative transfer results on FVL datasets. 
Without any supervised training on FVL, we observed that our model exhibit excellent transfer performance. We attribute this to: 
(1) our method does not need any prior processing which heavily rely on the data distribution, making them hard to transfer;
(2) the transformer-based network aggregates a richer context to focus on information that is more generalized to detect lane objects.

\section{Conclusion}
In this work, we present an end-to-end lane detector that directly outputs parameters of a lane shape model. 
The lane shape model reflects the road structures and camera state, enabling it to enhance the interpretability of the output parameters. 
The network built with transformer blocks efficiently learns global context to help infer occluded part and capture the long and thin structures especially nearby the horizon.
The whole method achieves state-of-the-art lane detection performance while requiring the least parameters and running time consumption. 
Meanwhile, our method adapts robustly to changes in datasets, making it easier to deploy on mobile devices and more reliable. 
It would be interesting to address complex and fine-grained lane detection tasks and introduce the tracking function in future work.

\section{Acknowledgement}
This work was supported in part by the National Key Research and Development Program of China under Grant 2016YFB1001001 and in part by the National Natural Science Foundation of China (61976170, 91648121, 61876112, 61603022).

\section{Appendix}
\label{appendix}
For reference, derivation for Eq.~\ref{tiltedcurve} is as follows. Given Eq.~\ref{poly3order}, according to the perspective projection, a pixel $\left(u, v\right)$ in the image plane projects on to the point $\left(X, Z\right)$ on the ground plane by:
\begin{equation}
\label{app-perspection}
X = u \times f_u \times Z;~~Z = \frac{H}{v \times f_v},
\end{equation}
where $f_u$ is the width of a pixel on the focal plane divided by the focal length, $f_v$ is the height of a pixel on the focal plane divided by the focal length, and $H$ is the camera height. Submitting Eq.~\ref{app-perspection} into Eq.~\ref{poly3order} and performing some polynomial simplification:
\begin{equation}
\label{app-imagecurve-ori}
u = \frac{k \times H^2}{f_u \times {f_v}^2 \times v^2} + \frac{m \times H}{f_u \times f_v \times v} + \frac{n}{f_u} + \frac{b \times f_v \times v}{f_u \times H},
\end{equation}
then combining parameters together we can get Eq.~\ref{imagecurve}.
Given the pitch angle $\phi$, the transformation between a tilted and an untitled camera is:
\begin{equation}
\label{app-coordtransf}
\begin{split}
u &= u', \\
\begin{bmatrix}
    f \\
    v
\end{bmatrix}
&= 
\begin{bmatrix}
    \cos \phi&\sin \phi \\
    -\sin \phi&\cos \phi
\end{bmatrix}
\begin{bmatrix}
    f' \\
    v'
\end{bmatrix}, \\
\begin{bmatrix}
    f' \\
    v'
\end{bmatrix}
&= 
\begin{bmatrix}
    \cos \phi&-\sin \phi \\
    \sin \phi&\cos \phi
\end{bmatrix}
\begin{bmatrix}
    f \\
    v
\end{bmatrix},
\end{split}
\end{equation}
where $\left(u, v, f\right)$ represents the location of a point in the untitled image plane, $f$ is the focal length which is in pixels, and $\left(u', v', f'\right)$ represents the pitch-transformed position of that point.
According to~Eq.~\ref{app-coordtransf}, we get:
\begin{equation}
\label{app-uvtransf}
v=\frac{v'-f \sin \phi}{\cos \phi}
\end{equation}
Submitting $u=u'$ and Eq.~\ref{app-uvtransf} into Eq.~\ref{imagecurve}, the curve function in the tilted image plane can be obtained with the form of Eq.~\ref{tiltedcurve}

{\small
\bibliographystyle{ieee_fullname}
\bibliography{ldtr}
}

\end{document}